\documentclass[sigconf]{acmart}

\usepackage{tabularx}
\usepackage{booktabs}
\usepackage{multirow}
\usepackage{multicol}
\usepackage{makecell}
\usepackage{pifont}
\usepackage{subcaption}
\usepackage{graphicx}
\usepackage{enumitem}
\usepackage{hyperref}
\usepackage{cleveref}
\usepackage{balance}
\copyrightyear{2025}
\acmYear{2025}
\setcopyright{acmlicensed}\acmConference[MSMA '25]{Proceedings of the 1st International Workshop on Multi-Sensorial Media and Applications}{October 27--28, 2025}{Dublin, Ireland}
\acmBooktitle{Proceedings of the 1st International Workshop on Multi-Sensorial Media and Applications (MSMA '25), October 27--28, 2025, Dublin, Ireland}
\acmDOI{10.1145/3728485.3759170}
\acmISBN{979-8-4007-1842-7/2025/10}
\begin{document}

\title{TacCompress: A Benchmark for Multi-Point Tactile Data Compression in Dexterous Hand}


\author{Yan Zhao}
\affiliation{%
  \institution{Shanghai Jiao Tong University}
  \city{Shanghai}
  \country{China}}
\email{zhaoyanzy@sjtu.edu.cn}

\author{Yang Li}
\affiliation{%
  \institution{PaXini Tech}
  \city{Shenzhen}
  \country{China}}
\email{liyang@paxini.com}

\author{Zhengxue Cheng}
\authornote{Corresponding author.}
\affiliation{%
  \institution{Shanghai Jiao Tong University}
  \city{Shanghai}
  \country{China}}
\email{zxcheng@sjtu.edu.cn}

\author{Hengdi Zhang}
\affiliation{%
  \institution{PaXini Tech}
  \city{Shenzhen}
  \country{China}}
\email{zhanghd@paxini.com}

\author{Li Song}
\affiliation{%
  \institution{Shanghai Jiao Tong University}
  \city{Shanghai}
  \country{China}}
\email{song\_li@sjtu.edu.cn}

\renewcommand{\shortauthors}{Yan Zhao, Yang Li, Zhengxue Cheng, Hengdi Zhang, and Li Song}
%% No italics, no superscripts
%% Use footnote or author note to identify equal contribution and/or contact author info

\begin{abstract}
Though robotic dexterous manipulation has progressed substantially recently, challenges like in-hand occlusion still necessitate fine-grained tactile perception, leading to the integration of more tactile sensors into robotic hands. Consequently, the increased data volume imposes substantial bandwidth pressure on signal transmission from the hand's controller. However, the acquisition and compression of multi-point tactile signals based on the dexterous hands' physical structures have not been thoroughly explored. In this paper, our contributions are twofold. First, we introduce a Multi-Point Tactile Dataset for Dexterous Hand Grasping (Dex-MPTD). This dataset captures tactile signals from multiple contact sensors across various objects and grasping poses, offering a comprehensive benchmark for advancing dexterous robotic manipulation research. Second, we investigate both lossless and lossy compression on Dex-MPTD by converting tactile data into images and applying six lossless and five lossy image codecs for efficient compression. Experimental results demonstrate that tactile data can be losslessly compressed to as low as 0.0364 bits per sub-sample (bpss), achieving approximately 200$\times$ compression ratio compared to the raw tactile data. Efficient lossy compressors like HM and VTM can achieve about 1000$\times$ data reductions while preserving acceptable data fidelity. The exploration of lossy compression also reveals that screen-content-targeted coding tools outperform general-purpose codecs in compressing tactile data.
\end{abstract}





\maketitle

\section{Introduction}

Tactile perception is fundamental to human sensing, enabling the recognition of object properties such as shape, texture, and temperature, while also providing essential feedback for precise manipulations \cite{bremner2017development}. In robotic systems, dexterous hands have been developed with increasing numbers of joints and contact points to mimic this capability, allowing more complex grasping strategies and adaptive interactions with diverse objects. 

\begin{figure}[!tb]
  \includegraphics[width=1.0\linewidth]{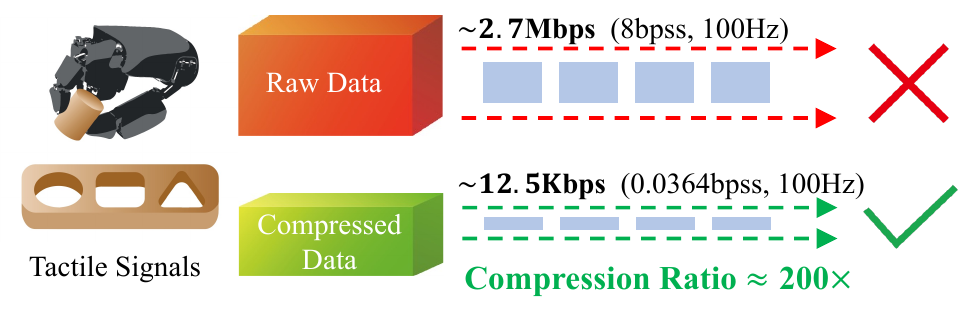}
  \caption{For a dexterous hand, at a frequency of 100Hz, the raw tactile data occupies a bandwidth of about 2.7 Mbits per second ($100~\text{Hz} \times1140~\text{units} \times3~\text{axis} \times8~\text{bpss}$), while after data format conversion and applying lossless compression, the required bandwidth drops to 12.5 Kbits per second ($100~\text{Hz}\times1140~\text{units}\times3~\text{axis}\times0.0364~\text{bpss}$), achieving more than 200$\times$ data volume reductions without information loss.}
  \Description{}
  \label{fig:teaser}
\end{figure}

To support research in robotic manipulation and perception, numerous tactile datasets have been introduced over the past decade, spanning applications such as object recognition, grasp planning, and material classification. However, few open-source datasets focus on multi-degree-of-freedom dexterous hands equipped with multi-point, array-based tactile sensors, which are crucial for studying fine-grained manipulation and whole-hand contact reasoning. At the same time, as modern robotic hands integrate more tactile sensors with higher sampling rates and spatial resolutions, the resulting data volume has grown substantially \cite{yuan2017gelsight}. This presents new challenges for real-time transmission, storage, and downstream processing. As illustrated in \cref{fig:teaser}, effective tactile data compression is vital not only for reducing communication bandwidth, but also for improving computational efficiency in latency-sensitive tasks \cite{8070953}.

\begin{figure*}[!tb]
  \includegraphics[width=\linewidth]{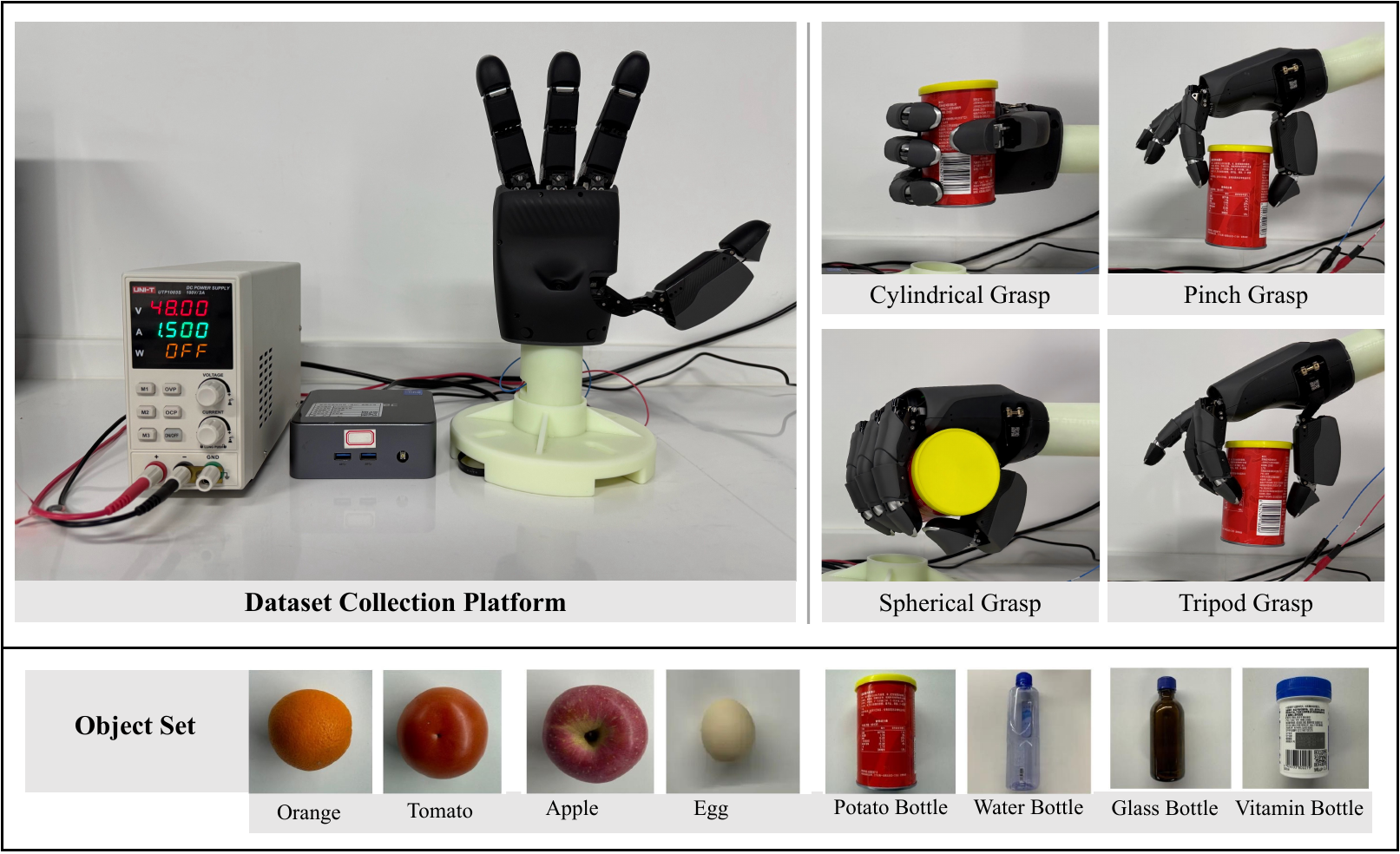}
  \caption{Dataset collection setup and object categories. Top left: experimental platform used for tactile data acquisition. Top right: four grasping poses including pinch, tripod, cylindrical, and spherical grasping. Bottom: eight objects used in the experiments, namely apple, egg, orange, tomato, potato, vitamin bottle, water bottle, and glass bottle.} 
  \Description{}
  \label{fig:platform}
\end{figure*}

Various approaches have investigated the compression of tactile data \cite{hollis2016compressed, shao2020compression, steinbach2018haptic, slepyan2024wavelet, lu2025cross, xu2024perception}. Classical methods based on signal processing (like dimensionality reduction and wavelet transforms) to reduce data redundancy but often struggle to fully preserve the complex features of tactile information \cite{watkins2018image}. In contrast, recent developments in neural network-based compression have shown promise for high-dimensional sensory data. Deep learning methods can learn compact latent representations in a data-driven manner, enabling efficient lossy or near-lossless compression while maintaining critical perceptual features \cite{tcm, l3c}. Compared to traditional codecs such as GZIP \cite{gzip}, WebP \cite{webp}, JPEG XL \cite{jpegxl}, VTM \cite{vtm}, and HM \cite{hm}, neural compression offers greater flexibility and adaptability, particularly in scenarios involving complex or irregular signal structures \cite{ma2019image}. These methods have been successfully applied in domains such as speech and image compression, and are increasingly being explored for tactile signals. 

However, most existing tactile compression methods are designed for single-sensor processing and do not fully account for the role of the dexterous hand’s physical structure in tactile perception \cite{ahanat2015tactile, 10710144, yang2021probabilistic}. In traditional studies, tactile perception is typically modeled using fingertip sensors alone, which provides only partial information and often proves insufficient for complex grasping and manipulation tasks \cite{bhirangi2023all}. In contrast, many real-world manipulation scenarios require coordinated, multi-point contact across different fingers or hand surfaces, resembling the way humans use multiple fingers to perceive and manipulate objects. With advancements in tactile data acquisition, emerging technologies like artificial skin and finger-like tactile sensors have expanded the potential for multi-point tactile sensing \cite{lepora2024future}. Consequently, there is a pressing need to design and optimize compression algorithms that are specifically tailored for multi-point tactile datasets. 

\begin{figure*}[t]
\includegraphics[width=1.0\linewidth]{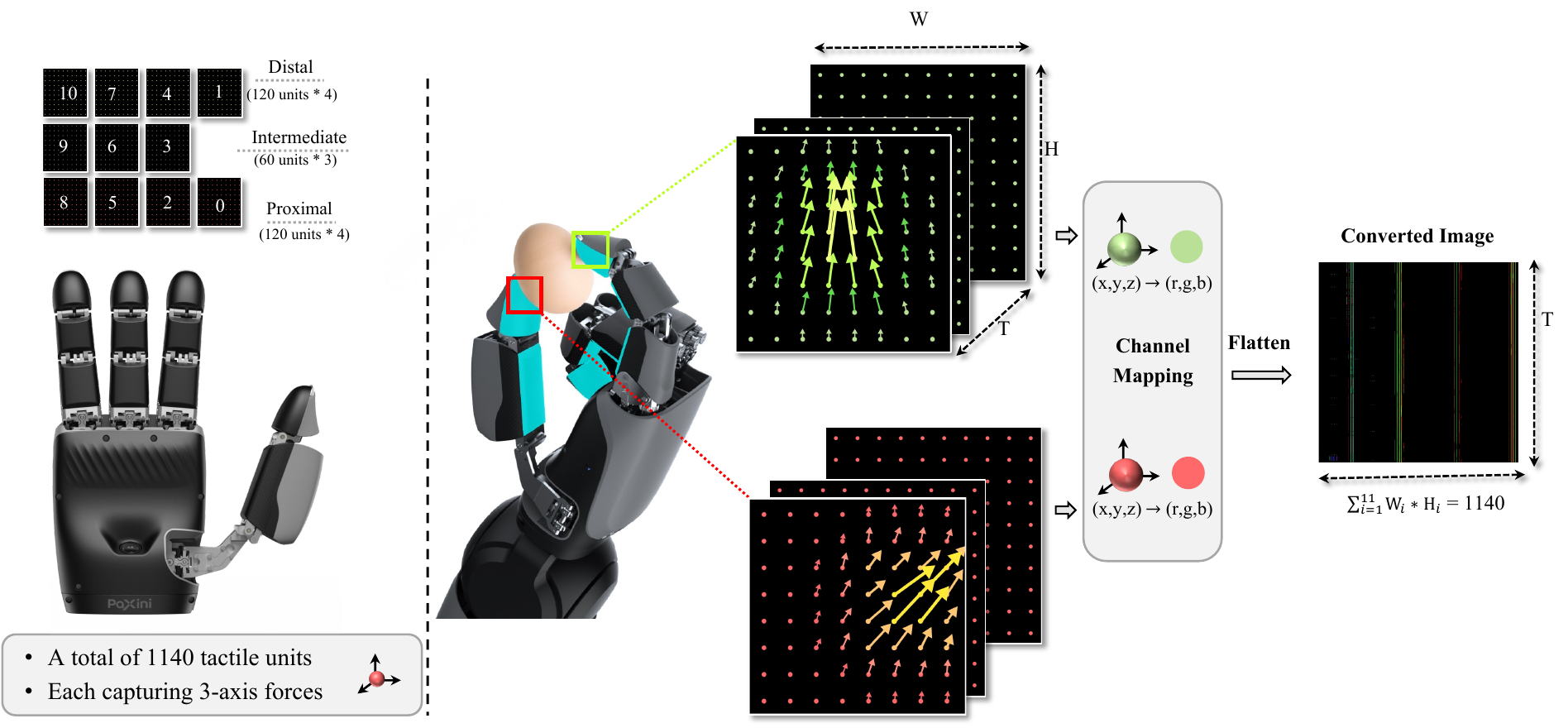}
    % \captionsetup{font=small}
  \caption{Illustration of the conversion method for transforming force data into images. A dexterous hand equipped with 11 sensors and 1,140 tactile units captures 3-axis force measurements at each contact point. The three-axis force vectors are encoded into RGB color channels, while spatially ordered sensor readings are arranged along the image width and temporally sequential data across timesteps define the image height.} 
  \label{fig:hand}
  \Description{}
\end{figure*}

To address these issues, we construct a multi-point tactile dataset based on dexterous hand grasping and thoroughly evaluate compression strategies on the proposed dataset. Generally speaking, our contributions can be summarized as:

\begin{itemize}
    \item We introduce a Multi-Point Tactile Dataset for Dexterous Hand Grasping (Dex-MPTD), which captures tactile signals from multiple contact points across the dexterous hand. By incorporating the spatial distribution of tactile feedback influenced by the hand's structure and grasping poses, this dataset provides a comprehensive benchmark designed to advance research in dexterous robotic manipulation. 
    \item We convert the collected tactile signals to images and conduct a comprehensive evaluation of compression performance. Six lossless and five lossy image compressors, including both traditional and neural-based methods, are compared for fairness and coverage. Among them, screen content coding tools demonstrate notable effectiveness.
    \item Experimental results show that lossless compression can reduce data volume to 0.0364 bpss (about 200$\times$ data reduction), significantly easing transmission and storage demands. Furthermore, efficient lossy compression achieves up to $1000\times$ reduction while maintaining acceptable perceptual quality. Application tests on an object classification task further validate the practicality of our compression strategy.
    
    % Highly efficient lossy compression further achieves about 1000$\times$ data reduction while maintaining acceptable distortion levels. Applications in object classification task also demonstrates the effectiveness of our proposed compression method.
    
    % \item Our results have indicated that we can compress the data lossless lower up to 0.0364bpsp, which greatly release the burden to transmit the tactile data.
\end{itemize}

\section{Related Work}
This section provides an overview of recent tactile datasets and corresponding tactile data compression methods, highlighting the distinctions between our approach and existing works.

\subsection{Tactile Dataset}
Tactile datasets are foundational to research in robotic perception, grasping, and manipulation. Over the past decade, a variety of datasets have been developed to support tasks such as object recognition, grasp planning, and contact modeling. As for the grasp-related tactile sensing, the STAG dataset \cite{sundaram2019learning} captures 135000 frames of full-hand pressure maps using a Velostat-based glove with 548 sensors arranged in a 32×32 grid, enabling fine-grained analysis of hand–object contact. The VTG dataset \cite{10710144} focuses on tripod grasp mechanics by equipping a tri-finger manipulator with 4×4 uSkin arrays on each joint, recording synchronized joint angles, 3-axis force data, and RGB-D point clouds across 18 textured objects. The Poselt dataset \cite{9981562} includes 1840 grasp trials spanning 26 objects and 16 grasp types, collected via a 6-DoF robotic arm with RGB-D cameras, per-finger tactile pads, and a wrist-mounted force/torque sensor.

Beyond grasping, other datasets focus on material perception and interactive behavior. For example, the PFN-VT dataset \cite{takahashi2019deep} collects tactile responses from 25 materials using a uSkin-equipped robotic arm to estimate surface properties like roughness and softness. The tactile signature dataset \cite{cepria2021dataset}, one of the largest open-access tactile datasets, captures hand tactile data from 22 subjects performing 21 daily activities. The 3D object dataset \cite{garcia2017multi} integrates Kinect V2 and other sensors to support robotic grasping and object recognition research. The multimodal tactile dataset \cite{lima2023multimodal} combines pressure, gravity, angular velocity, and magnetic field data to enhance tactile texture classification through machine learning. The Baxter tactile dataset \cite{tatiya2023transferring} collects feedback across multiple objects and grasping poses for robot perception studies. The Ego pressure dataset provides pressure intensity data for each contact point, aiding research on tactile contact and pressure interaction \cite{zhao2024egopressure}. 

However, despite these advances, few open-source datasets focus on multi-degree-of-freedom dexterous hands equipped with multi-point array-based tactile sensors. To bridge this gap, we present a multi-point tactile dataset collected using a multi-DOF dexterous hand. The dataset records tactile signals from numerous contact points across diverse objects and grasping poses, providing a benchmark for research in dexterous robotic manipulation.

\subsection{Tactile Compression Methods}

As robotic tactile systems evolve, the volume and frequency of sensor data continue to grow, posing increasing challenges for transmission and on-board storage, particularly given the limited bandwidth of micro-control units (MCUs) embedded in robotic hands. Without compression, tactile signals may exceed communication capacities, impairing system responsiveness and degrading control performance. This makes effective compression essential for maintaining real-time operation and reducing memory overhead.

To address these issues, recent studies have explored efficient tactile data compression techniques aimed at improving responsiveness and reducing storage overhead. For example, \citet{hollis2016compressed} apply compressed sensing to tactile sensor arrays, exploiting signal sparsity to minimize data volume. \citet{steinbach2018haptic} introduce a codec framework designed to support low-latency and high-fidelity tactile transmission. \citet{shao2020compression} demonstrate that mechanical wave propagation during dynamic contact enables compact tactile signal encoding. \citet{slepyan2024wavelet} leverage wavelet transforms to enhance signal sparsity, facilitating compression while preserving perceptual quality. \citet{xu2024perception} propose a kinesthetic coding scheme that utilizes dead-zone quantization and segmented linear prediction. Meanwhile, \citet{lu2025cross} present a cross-modal compression framework that incorporates visual semantics and multi-head attention mechanisms to improve tactile encoding performance.

However, existing tactile compression methods focus  primarily on single-sensor processing and overlook the role of dexterous hand structure in tactile perception. Therefore, we propose a data conversion strategy that transforms multi-point tactile signals into images, enabling the use of efficient image compression techniques to better exploit the redundancies of tactile data.

\section{Dataset Collection}

This section describes the data collection setup and acquisition process for the Dex-MPTD dataset, as illustrated in \cref{fig:platform}.

\subsection{Data Collection Platform}

While numerous tactile datasets exist, few of them are based on multi-degree-of-freedom dexterous hands equipped with multiple tactile sensors. These existing datasets often fall short when it comes to complex tasks involving high-precision, multi-tasking, or multi-object manipulation, which limits their applications in advancing intelligent robots for multi-degree-of-freedom manipulation. Therefore, we present a tactile dataset built on a multi-degree-of-freedom dexterous hand \emph{DexH13} developed by PaXiniTech \cite{paxiniDexH13}, as illustrated in \cref{fig:platform}. This dataset involves four different grasping poses (pinch, tripod, cylindrical, and spherical grasping), providing researchers with diverse experimental conditions.

As shown in \cref{fig:platform}, the data collection platform consists of PaxiniTech's dexterous hand, a 48-volt DC power supply, and a hand controller. The dexterous hand has four flexible fingers, each driven by a rotary motor, enabling actions like grasping, gripping, pinching, pressing, and finger opening and closing. This design mimics human hand gestures, allowing a more accurate presentation of complex tactile scenarios. By precisely controlling the four fingers, we can capture multi-point tactile data across various movement modes. As shown in \cref{fig:hand}, the dexterous hand has a total of 11 sensors across its four fingers. Since human fingers are most sensitive at the distal and proximal ends, each distal and proximal sensor on the dexterous hand is equipped with 120 tactile units, while the intermediate sensors have 60 units each. Hence, the hand has a total of 1140 tactile units, each capturing 3-dimensional spatial signals.

\subsection{Data collection Method}

We construct a multi-point tactile dataset based on PaxiniTech's multi-degree-of-freedom dexterous hand. The dataset includes four grasping poses: \textit{pinch}, \textit{tripod}, \textit{cylindrical}, and \textit{spherical} grasping. For each grasping pose, experiments are conducted on eight different objects: \textit{apple}, \textit{egg}, \textit{orange}, \textit{tomato}, \textit{potato}, \textit{vitamin bottle}, \textit{water bottle}, and \textit{glass bottle}, as illustrated in \cref{fig:platform}. To enhance data diversity and stability, each combination of object and grasping pose is repeated ten times for data collection.

Specifically, the object grasping experiments can be divided into two main steps: grasping and releasing.

\begin{itemize}
    \item \textit{Grasping}: For each grasping pose, the dexterous hand grasps the object and vertically lifts it by 0.3 meters to ensure stable grasping. The data collection starts 10 seconds before the grasping action to ensure data accuracy, with each tactile unit capturing 3-dimensional force information.

    \item \textit{Releasing}: After lifting the object to a designated position, the dexterous hand maintains a static hold for 15 seconds to record steady-state tactile feedback. The object is then released, marking the end of data collection. This phase primarily captures tactile variations during the transition from grasping to release, contributing to a more comprehensive understanding of contact dynamics.

\end{itemize}

\section{Data Compression Evaluation}

This section presents a series of evaluations on the proposed multi-point tactile dataset. We first convert the raw tactile data into RGB images, then apply a range of lossless and lossy image compression algorithms to assess their efficiency. In addition, we validate the utility of the compressed data through an object classification task, demonstrating its potential for downstream robotic applications.

\begin{table}[!t]
    \centering
    \footnotesize
     \renewcommand{\arraystretch}{1.15}
    \setlength{\tabcolsep}{4pt} % 调整列间距
    % \captionsetup{font=small}
    \caption{Lossless compression performance (bpss) across objects and grasping poses. The uncompressed raw data by default has 8 bits per sub-sample, then compression ratio (CR) in the last row is calculated by 8.0 / bpss.}
    \begin{tabularx}{1.0\columnwidth}{c|c|cccccc}
    \toprule
    % & & \multicolumn{6}{c}{\textbf{bits per sub-sample (bpss)}} \\ 
    % \cmidrule[0.5pt]{3-8} % 只在2-8列上加分割线
    \multicolumn{2}{c|}{Settings} & \textbf{BPG} & \textbf{FLIF} & \textbf{JPEG-XL} & \textbf{WebP} & \textbf{GZIP} & \textbf{L3C}  \\
    \midrule
    % \cmidrule[0.5pt]{3-8}
    \multirow{8}{*}{\rotatebox{90}{\textbf{Object}}} & Orange & 0.0665 & 0.0491 & 0.0509 & 0.0396 & 0.0622 & 0.0616 \\
    & Tomato & 0.0617 & 0.0441 & 0.0430 & 0.0345 & 0.0585 & 0.0575 \\
    & Apple  & 0.0756 & 0.0486 & 0.0528 & 0.0452 & 0.0706 & 0.0669 \\
    & Egg    & 0.0402 & 0.0290 & 0.0308 & 0.0236 & 0.0403 & 0.0434 \\
    & Potato Bottle & 0.0650 & 0.0453 & 0.0438 & 0.0355 & 0.0583 & 0.0580 \\
    & Water Bottle  & 0.0570 & 0.0438 & 0.0405 & 0.0321 & 0.0567 & 0.0571 \\
    & Glass  & 0.0635 & 0.0448 & 0.0467 & 0.0411 & 0.0625 & 0.0613 \\
    & Vitamin Bottle  & 0.0662 & 0.0481 & 0.0459 & 0.0398 & 0.0621 & 0.0608 \\
    \midrule
    \multirow{4}{*}{\rotatebox{90}{\textbf{Grasp}}} & Pinch & 0.0567 & 0.0411 & 0.0407 & 0.0339 & 0.0537 & 0.0550\\
    & Tripod & 0.0570 & 0.0438 & 0.0409 & 0.0339 & 0.0552 & 0.0560\\
    & Cylindrical & 0.0730 & 0.0465 & 0.0517 & 0.0416 & 0.0655 & 0.0642\\
    & Spherical & 0.0601 & 0.0443 & 0.0439 & 0.0357 & 0.0600 & 0.0570\\
    \midrule
    \multicolumn{2}{c|}{\textbf{Average bpss}}  & \textbf{0.0619} & \textbf{0.0441} & \textbf{0.0443} & \textbf{0.0364} & \textbf{0.0589} & \textbf{0.0583} \\
    \midrule
    \multicolumn{2}{c|}{\textbf{CR}}  & \textbf{129$\times$} & \textbf{181$\times$} & \textbf{181$\times$} & \textbf{220$\times$} & \textbf{136$\times$} & \textbf{137$\times$} \\    
   \bottomrule
    \end{tabularx}
    \label{tab:lossless}
\end{table}

\subsection{Preprocessing for Efficient Compression}
\label{sec:preprocess}

To enable effective compression and downstream analysis, we represent tactile signals in RGB image format. This choice is motivated by two considerations. First, each tactile sensing unit produces three-axis force measurements, which map naturally to the red, green, and blue channels of an image. Second, the image domain provides a well-established suite of tools for compression, classification, and clustering, with proven performance and broad applicability. As illustrated in \cref{fig:hand}, the DexH13 dexterous hand integrates 11 sensor arrays, containing 1140 individual sensing units. Each unit captures three-dimensional spatial stress signals relative to its local coordinate system, measuring both parallel and perpendicular components with respect to the local sensor surface. Through real-time fusion of these localized measurements with the hand's joint angle data, high-resolution contact stress distributions across the entire manipulator can be reconstructed.

To convert tactile data into an image format, we first collect a sequence of tactile signals over a duration of $T$ frames. Then we concatenate them along the temporal dimension, forming a matrix of size $(1140, T, 3)$. The three force components are then directly mapped to the R, G, and B channels, resulting in an RGB image with a resolution of $1140\times T$. This approach not only preserves all tactile information but also enables the removal of temporal and sensor redundancies during subsequent data compression process. Furthermore, with such data format conversion, we can also apply computer vision techniques to process tactile signals, such as clustering \cite{clustersurvey} and classification \cite{clssurvey}.

\begin{figure}
  \includegraphics[width=1.0\linewidth]{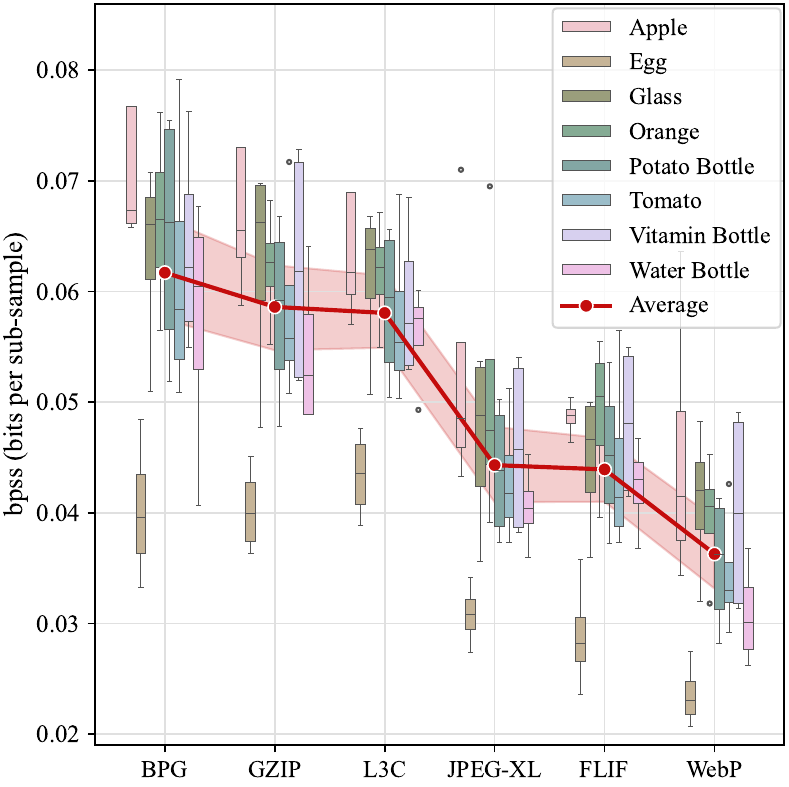}
  % \captionsetup{font=small}
  \caption{Lossless compression performance. The x-axis shows various compressors. The y-axis presents their compression efficiency (bpss). The boxes indicate performance variation across different grasping poses for each object, while the red line indicates each compressor's average performance across all objects.} 
  \label{fig:lossless}
\end{figure}

\begin{figure}
  \includegraphics[width=1.0\linewidth]{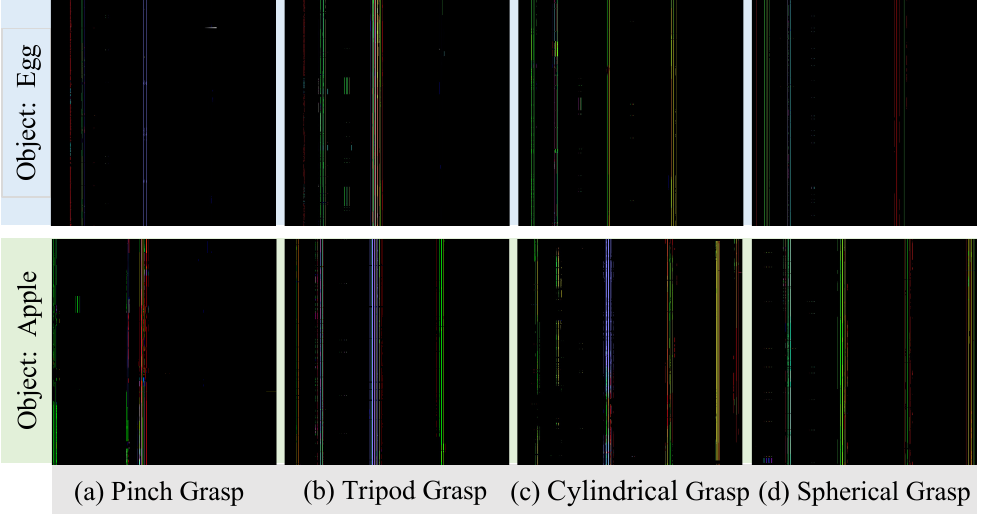}
  % \captionsetup{font=small}
  \caption{Visualization of images converted from tactile signals. The first two rows correspond to grasping egg and apple, respectively. The four columns represent four grasping poses. Eggs produce smoother tactile images due to their regular shape and light weight, while apples yield noisier patterns owing to irregular geometry and greater mass.} 
  \label{fig:images}
\end{figure}

\subsection{Compression Methods}

This paper applies both lossless and lossy compression algorithms to tactile signal images, comparing their compression efficiency to offer a comprehensive evaluation.

\begin{figure*}[!tbp]
    \centering
    \begin{subfigure}{0.49\linewidth}
        \centering
        \includegraphics[width=\linewidth]{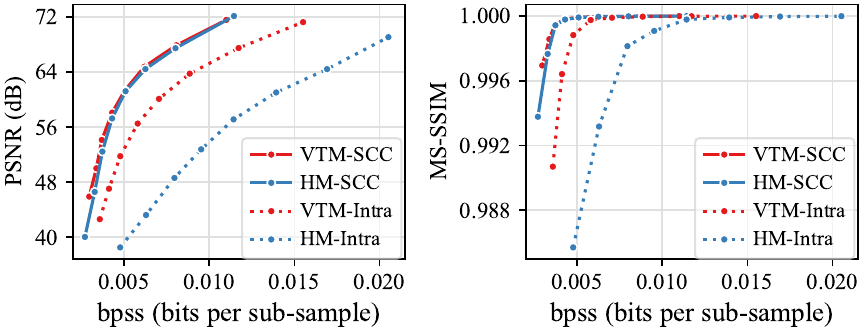}
        \caption{\small Object: Egg.}
    \end{subfigure}
    \hfill
    \begin{subfigure}{0.49\linewidth}
        \centering
        \includegraphics[width=\linewidth]{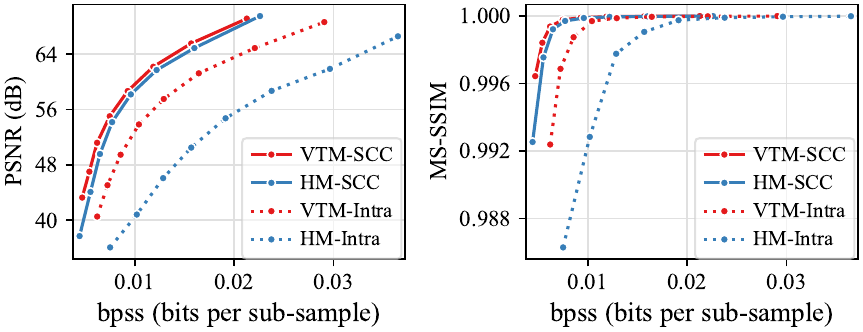}
        \caption{\small Object: Apple.}
    \end{subfigure}\\
    \vspace{0.3cm}
    \begin{subfigure}{0.49\linewidth}
        \centering
        \includegraphics[width=\linewidth]{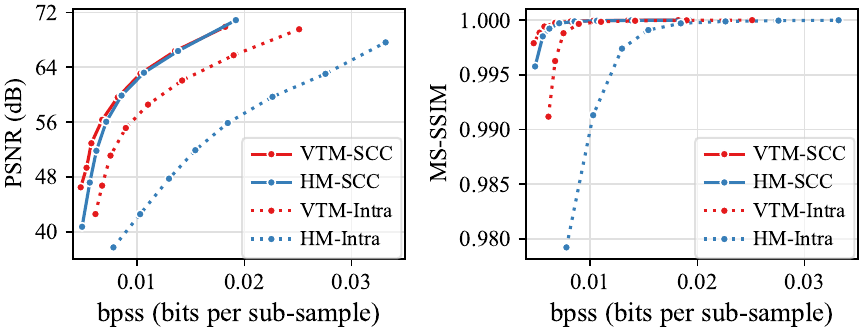}
        \caption{\small Object: Water bottle.}
        \label{fig:ablation-structure}
    \end{subfigure}
    \hfill
    \begin{subfigure}{0.49\linewidth}
        \centering
        \includegraphics[width=\linewidth]{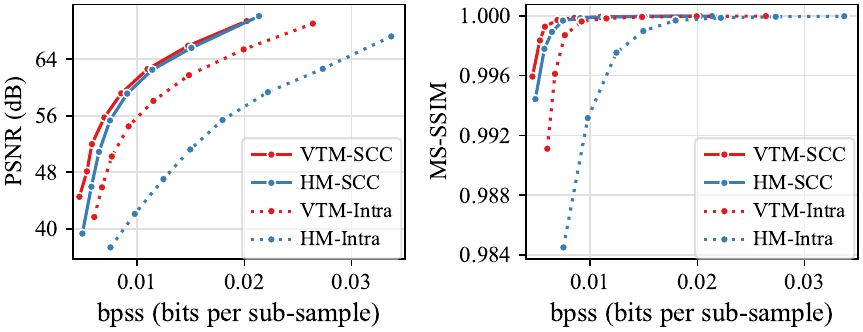}
        \caption{\small Object: Vitamin bottle.}
        \label{fig:ablation-scale}
    \end{subfigure}
    \Description{}
    \caption{Lossy compression performance with and without SCC tools in VTM and HM. Each subfigure corresponds to an object.}
    \label{fig:lossy-scc-objects}
    
\end{figure*}

\begin{figure*}[!tbp]
    \centering
    \includegraphics[width=0.86\linewidth]{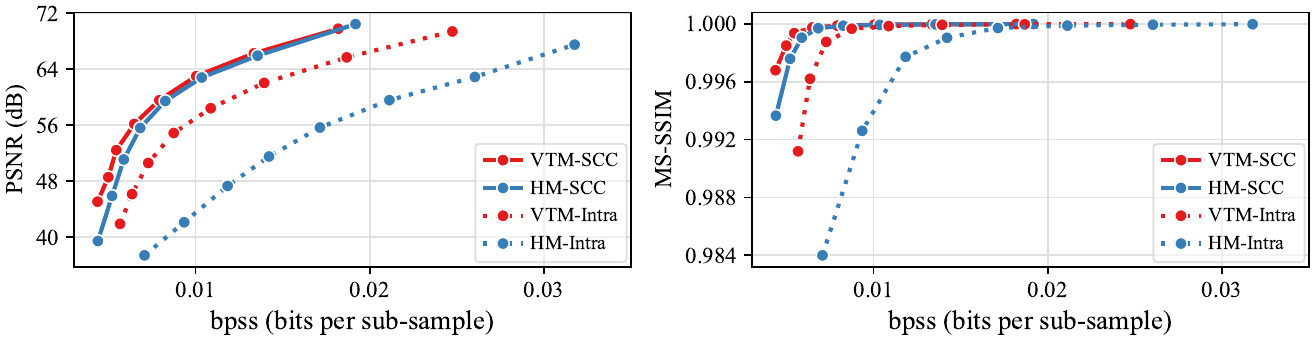}
    \caption{Average lossy compression performance across all objects, comparing intra coding with SCC in VTM and HM.}
    \label{fig:lossy-scc-all}
\end{figure*}

Specifically, we employ six lossless compression methods: the neural network-based L3C \cite{l3c} and five traditional non-neural techniques, namely BPG \cite{bpg}, FLIF \cite{flif}, JPEG-XL \cite{jpegxl}, WebP \cite{webp}, and GZIP \cite{gzip}, to thoroughly assess the compression efficiency of the converted signals. These methods are selected to reflect a diverse set of compression strategies, including entropy coding, predictive coding, and learned representations. For lossy compression, we examine the rate-distortion (RD) performance of five widely used image codecs: the intra-frame coding configurations of HM \cite{hm} and VTM \cite{tcm}, two well-established traditional codecs (JPEG2000 \cite{jpeg2000} and JPEG-XL \cite{jpegxl}), and the neural network-based TCM \cite{tcm}. For L3C and TCM, we use publicly available pre-trained models to ensure reproducibility and fair comparison.

Moreover, as visualized in \cref{fig:hand}, the tactile images exhibit structured patterns with sharp transitions and localized high-frequency content, characteristics commonly seen in screen content images. Motivated by this similarity, we further evaluate the performance of screen content coding (SCC) tools built into HM and VTM \cite{scc-hm, scc-vtm}, which are specifically optimized for such data structures and may offer superior compression performance in this domain.

\subsection{Lossless Compression Evaluation}

\cref{tab:lossless} and \cref{fig:lossless} illustrate the lossless compression performance of various algorithms for tactile data, measured in bits per sub-sample (bpss), with lower bpss indicating better compression performance. For uncompressed raw data, each sub-sample is stored using 8 bits. The upper part of \cref{tab:lossless} summarizes the average performance per algorithm and object, while the bottom part summarizes the average performance per algorithm and grasping pose. In \cref{fig:lossless}, the x-axis shows compression methods, and the y-axis represents compression efficiency. The boxes indicate performance variation across different poses (cylindrical, spherical, pinch, tripod) when grasping an object. The red line represents the average performance for each compression method across all objects.

It can be seen from \cref{tab:lossless} that all the lossless compressors exhibit satisfactory compression efficiency, with average bpss ranging from 0.0364 to 0.0619, corresponding to compression ratios of 100 to 200 times compared to uncompressed data with bpss of 8.0. While BPG, GZIP, and L3C offer moderate compression performance (0.0583 to 0.0619 bpss), JPEG-XL, FLIF, and WebP achieve notably better results, with average bpss of 0.0441 and 0.0364, respectively. This advantage is attributable to their underlying compression mechanisms. FLIF, for instance, employs the MANIAC algorithm \cite{flif}, which adaptively learns local data distributions and adjusts its encoding strategy accordingly. WebP lossless compression, on the other hand, builds on LZ77 \cite{lz77} and Huffman coding, leveraging long-match string referencing to exploit spatial redundancy. This makes it particularly well-suited for tactile images, which often exhibit repetitive spatial patterns.

\begin{figure*}[!t]
    \centering
    \begin{subfigure}{0.49\linewidth}
        \centering
        \includegraphics[width=\linewidth]{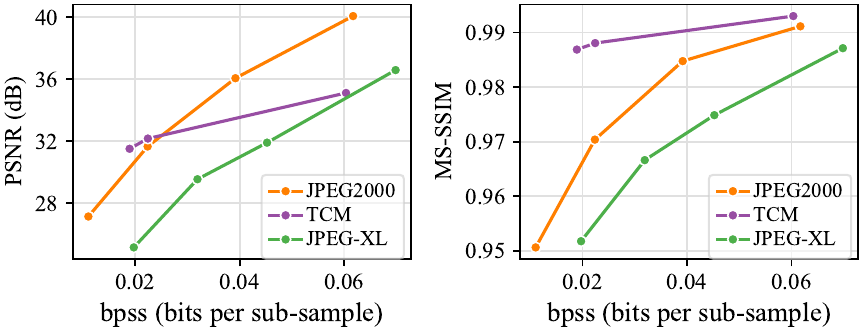}
        \caption{\small Object: Potato bottle.}
    \end{subfigure}
    \hfill
    \begin{subfigure}{0.49\linewidth}
        \centering
        \includegraphics[width=\linewidth]{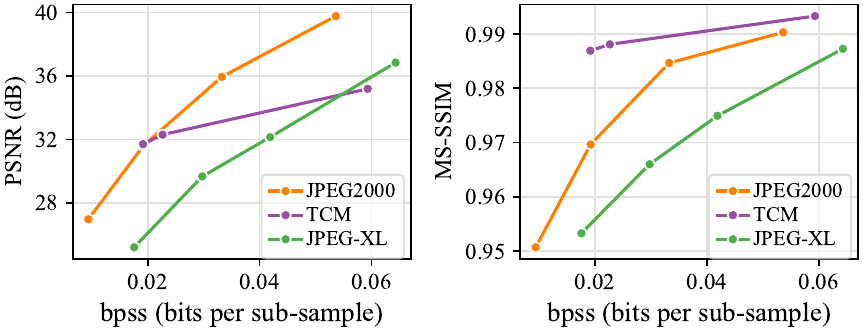}
        \caption{\small Object: Tomato.}
    \end{subfigure}\\
    \vspace{0.3cm}
    \begin{subfigure}{0.49\linewidth}
        \centering
        \includegraphics[width=\linewidth]{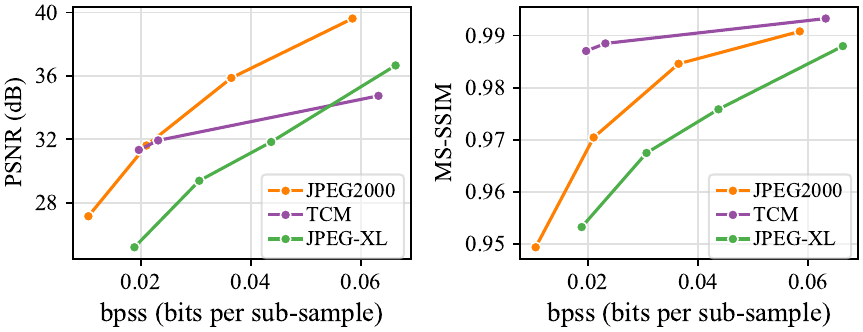}
        \caption{\small Object: Orange.}
    \end{subfigure}
    \hfill
    \begin{subfigure}{0.49\linewidth}
        \centering
        \includegraphics[width=\linewidth]{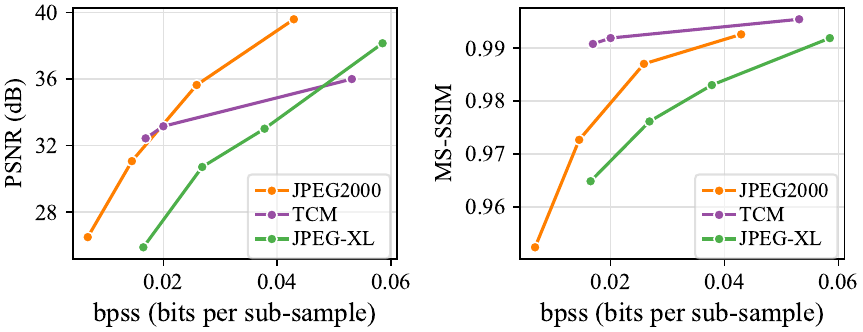}
        \caption{\small Object: Glass.}
    \end{subfigure}
    % \captionsetup{font=small}
    \caption{Lossy compression performance using general-purpose image codecs. Each sub-figure corresponds to an object.}
    \label{fig:lossy-gen-objects}
    
\end{figure*}

\begin{figure*}
    \centering
    \includegraphics[width=0.86\linewidth]{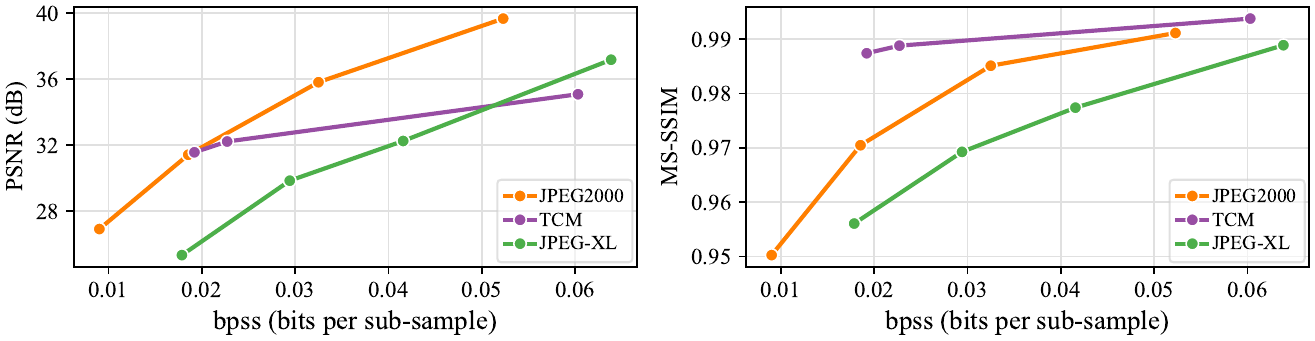}
    \caption{Average lossy compression performance across all the objects using general-purpose image compressors.}
    \label{fig:lossy-gen-all}
\end{figure*}

From the lower part of \cref{tab:lossless}, it can be observed that the cylindrical grasping presents higher bpss because the tactile is difficult to compress, which implies cylindrical grasping affords greater force pressure. Meanwhile, the precision grasping, including pinch and tripod grasping poses, offers less force. This observation aligns with the typical force distribution in human grasping behavior. Furthermore, across all compressors, tactile signals from grasping eggs achieve the highest compression efficiency, while those from grasping apples and bottles are notably harder to compress. \cref{fig:images} visualizes the tactile images generated by these two objects under different grasping poses. Eggs are lightweight and geometrically regular, hence produce smoother and more structured tactile images, making compression more efficient. In contrast, apples' irregular shapes and heavier weights introduce more noise and artifacts in the converted tactile images, increasing the compression difficulty.

\subsection{Lossy Compression Evaluation}

\cref{fig:lossy-scc-objects}, \cref{fig:lossy-scc-all}, \cref{fig:lossy-gen-objects}, and \cref{fig:lossy-gen-all} illustrate the rate-distortion performance when applying lossy compression, using PSNR \cite{psnr} and MS-SSIM \cite{ssim} as the distortion metrics. As shown in \cref{fig:images}, tactile images exhibit several characteristics commonly found in screen content, such as sharp contours, limited color variation, and repetitive spatial patterns. These properties make screen content coding (SCC) tools particularly suitable for compressing such data. Therefore, we apply the SCC tools in HM and VTM \cite{sccoverview}, and compare their performance against the standard intra coding modes, as shown in \cref{fig:lossy-scc-objects} and \cref{fig:lossy-scc-all}. Notably, HM and VTM maintain reconstruction quality of no less than 38 dB at extremely low bit rates ($<$0.01 bpss, approximately 800 $\times$ compression ratios) even using normal intra encoding mode, demonstrating their effectiveness for this task. Similar to lossless compression, lightweight and geometrically regular objects, such as eggs and water bottles, achieve better rate-distortion performance than objects like apples and vitamin bottles, as shown in \cref{fig:lossy-scc-objects}. Meanwhile, it can be observed that for the same codec, enabling SCC tools largely improves the compression efficiency of tactile images, achieving 30.67\% and 56.42\% BD-Rate \cite{bjontegaard} gains for VTM and HM, respectively.

% Given the similarity of the tactile images to screen content, we first apply the state-of-the-art screen content coding (SCC) tools \cite{sccoverview} in HM and VTM, comparing their performance against standard intra coding mode, as shown in \cref{fig:lossy-scc-objects} and \cref{fig:lossy-scc-all}.

We also evaluate other lossy image codecs like JPEG2000, JPEG-XL, and TCM, as shown in \cref{fig:lossy-gen-objects} and \cref{fig:lossy-gen-all}, where the prior demonstrates distinct objects' compression performance, and the latter shows the average performance across all the objects. The three methods are all general-purpose codecs designed for all types of images and deliver moderate compression ratios for our tactile signals. They exhibit inferior encoding performance compared to HM and VTM in both PSNR and MS-SSIM. Neural-based compressors like TCM and L3C often struggle with generalization capability and perform sub-optimally on such out-of-domain data. However, while TCM's PSNR performance is inferior to JPEG2000, it excels in terms of the MS-SSIM metric.

\subsection{Applications on Downstream Tasks}

\begin{figure}[!tbp]
\includegraphics[width=1.0\linewidth]{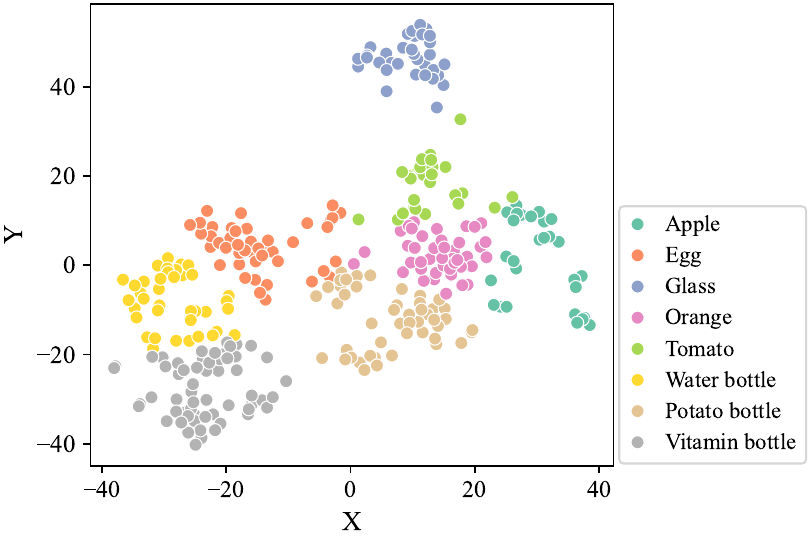}
  % \captionsetup{font=small}
  \caption{Clustering results on the proposed Dex-MPTD dataset using t-SNE projection and K-means.} 
  \label{fig:cluster}
\end{figure}

To validate the effectiveness of the proposed dataset for downstream tasks, we conduct object classification experiments on the generated tactile images. First, we perform a clustering analysis to verify the separability of the generated data. The uncompressed images are projected to 2D space using t-SNE \cite{tsne}, followed by K-means clustering \cite{kmeans} with $k$ set to the number of object categories. As illustrated in \cref{fig:cluster}, the tactile images of each object form distinct clusters, indicating that object-specific features are preserved in the dataset and it is feasible to perform object classification.

Then, we simply employ some classical machine learning (ML) methods, including linear regression (LR) \cite{linregression}, support vector machine (SVM) \cite{svm}, random forest (RF) \cite{randomforest}, and K-nearest neighboring (K-NN) \cite{knn}, to classify the corresponding object of each tactile image. To ensure a balanced evaluation, 70\% of the images from each object are randomly selected as the training set, while the remaining serve as the test set. The images are normalized and flattened into vectors as inputs to the classification models. As presented in \cref{tab:accuracy}, when classifying uncompressed raw images, all these methods achieve at least 70\% accuracy, among which SVM and LR perform the best, with accuracies of 81.25\% and 79.17\%, respectively. LR's satisfactory performance suggests that tactile image data exhibit relatively well-structured feature distributions. The superior performance of SVM can be attributed to its ability to handle high-dimensional feature spaces and effectively separate complex patterns, making it well-suited for tactile image classification.

\begin{table}[!t]
    \centering
    \footnotesize
     \renewcommand{\arraystretch}{1.15}
    \setlength{\tabcolsep}{7.5pt} % 调整列间距
    % \captionsetup{font=small}
    \caption{Object classification accuracy on raw data and VTM-SCC compressed dats using different classifiers.}
    \begin{tabularx}{\columnwidth}{cc|cccc}
    \toprule
    \textbf{Data} & \textbf{bpss} & \textbf{SVM} \cite{svm} & \textbf{RF} \cite{randomforest} & \textbf{K-NN} \cite{knn} & \textbf{LR} \cite{linregression} \\
    \midrule
    \textbf{Raw Data} & 8.0000 & 81.25\% & 76.04\% & 70.93\% & 79.17\% \\
    \midrule
    \multirow{3}{*}{\textbf{VTM-SCC}} & 0.0160 & 80.21\% & 72.92\% & 65.62\% & 76.04\% \\
                 & 0.0110 & 77.08\% & 71.88\% & 64.58\% & 73.96\% \\
                 & 0.0067 & 72.92\% & 66.67\% & 61.45\% & 70.83\% \\
   \bottomrule
    \end{tabularx}
    \label{tab:accuracy}
\end{table}

Further, we also analyze the impact of compression on object classification using VTM-SCC, the most efficient lossy compressor. As shown in \cref{tab:accuracy}, compared to performing object classification on raw data, VTM reduces the data volume by a factor of 1200 (i.e., 0.0067 bpss), while still maintaining high classification accuracy (over 70\% with SVM and LR). When the data volume is compressed by a factor of 500 (i.e., 0.016 bpss), the resulting distortion has minimal impact on the classification task, particularly for SVM (with only 1\% accuracy loss compared to classifying the uncompressed raw data). These results demonstrate the robustness of the proposed dataset for downstream tasks and highlight the effectiveness of our compression methods in reducing tactile data size while retaining essential information for accurate classification.

\section{Conclusion}

To achieve high transmission and realtime processing for dexterous manipulation, in this paper we construct a first benchmark of multi-point tactile data compression. First, we introduce the Dex-MPTD, a multi-point tactile dataset for dexterous hand grasping, which captures tactile signals across various sensors, objects, and grasping poses. This dataset serves as a valuable benchmark for advancing research in tactile perception and robotic dexterous manipulation. Second, to address the challenge of high data bandwidth in multi-point tactile sensing, we extensively explore lossless compression techniques by converting tactile signals into images and leveraging modern image compression codecs. Experiments demonstrate that lossless compression can reduce tactile data to 0.0364 bits per sub-sample, achieving up to 200$\times$ compression ratio. Efficient lossy compressors achieve approximately 1000$\times$ data volume reduction while maintaining acceptable fidelity. Further, we investigate general-purpose and SCC-targeted lossy compression methods, revealing that screen content coding tools are effective for tactile data compression. Downstream applications have been discussed to demonstrate that the lossy compression methods only result to 1\% accuracy loss for classification task, compared to uncompressed data, which illustrates the efficiency of utilized compression methods. Our codes and dataset will be released upon acceptance.

\section{Acknowledgment}

This work was partly supported by the Paxini Tech Research Fund, NSFC62431015, Science and Technology Commission of Shanghai Municipality No.24511106200, the Fundamental Research Funds for the Central Universities, Shanghai Key Laboratory of Digital Media Processing and Transmission under Grant 22DZ2229005, 111 project BP0719010.

\bibliographystyle{ACM-Reference-Format}
\balance
\bibliography{reference}


\begin{thebibliography}{50}

%%% the Web address.

\ifx \showCODEN    \undefined \def \showCODEN     #1{\unskip}     \fi
\ifx \showISBNx    \undefined \def \showISBNx     #1{\unskip}     \fi
\ifx \showISBNxiii \undefined \def \showISBNxiii  #1{\unskip}     \fi
\ifx \showISSN     \undefined \def \showISSN      #1{\unskip}     \fi
\ifx \showLCCN     \undefined \def \showLCCN      #1{\unskip}     \fi
\ifx \shownote     \undefined \def \shownote      #1{#1}          \fi
\ifx \showarticletitle \undefined \def \showarticletitle #1{#1}   \fi
\ifx \showURL      \undefined \def \showURL       {\relax}        \fi
% invisible to TeX
\providecommand\bibfield[2]{#2}
\providecommand\bibinfo[2]{#2}
\providecommand\natexlab[1]{#1}
\providecommand\showeprint[2][]{arXiv:#2}

\bibitem[Ahanat et~al\mbox{.}(2015)]%
        {ahanat2015tactile}
\bibfield{author}{\bibinfo{person}{K Ahanat}, \bibinfo{person}{ACR Juan}, {and} \bibinfo{person}{P Veronique}.} \bibinfo{year}{2015}\natexlab{}.
\newblock \showarticletitle{Tactile sensing in dexterous robot hands-Review}.
\newblock \bibinfo{journal}{\emph{Rob. Auton. Syst}}  \bibinfo{volume}{74} (\bibinfo{year}{2015}), \bibinfo{pages}{195--220}.
\newblock


\bibitem[Bakurov et~al\mbox{.}(2022)]%
        {ssim}
\bibfield{author}{\bibinfo{person}{Illya Bakurov}, \bibinfo{person}{Marco Buzzelli}, \bibinfo{person}{Raimondo Schettini}, \bibinfo{person}{Mauro Castelli}, {and} \bibinfo{person}{Leonardo Vanneschi}.} \bibinfo{year}{2022}\natexlab{}.
\newblock \showarticletitle{Structural similarity index (SSIM) revisited: A data-driven approach}.
\newblock \bibinfo{journal}{\emph{Expert Systems with Applications}}  \bibinfo{volume}{189} (\bibinfo{year}{2022}), \bibinfo{pages}{116087}.
\newblock


\bibitem[Bellard(2014)]%
        {bpg}
\bibfield{author}{\bibinfo{person}{Fabrice Bellard}.} \bibinfo{year}{2014}\natexlab{}.
\newblock \bibinfo{title}{BPG Image format}.
\newblock \bibinfo{howpublished}{\url{https://bellard.org/bpg/}}.
\newblock
\newblock
\shownote{Accessed: 2025-02-28}.


\bibitem[Bhirangi et~al\mbox{.}(2023)]%
        {bhirangi2023all}
\bibfield{author}{\bibinfo{person}{Raunaq Bhirangi}, \bibinfo{person}{Abigail DeFranco}, \bibinfo{person}{Jacob Adkins}, \bibinfo{person}{Carmel Majidi}, \bibinfo{person}{Abhinav Gupta}, \bibinfo{person}{Tess Hellebrekers}, {and} \bibinfo{person}{Vikash Kumar}.} \bibinfo{year}{2023}\natexlab{}.
\newblock \showarticletitle{All the feels: A dexterous hand with large-area tactile sensing}.
\newblock \bibinfo{journal}{\emph{IEEE Robotics and Automation Letters}} \bibinfo{volume}{8}, \bibinfo{number}{12} (\bibinfo{year}{2023}), \bibinfo{pages}{8311--8318}.
\newblock


\bibitem[Bjontegaard(2001)]%
        {bjontegaard}
\bibfield{author}{\bibinfo{person}{Gisle Bjontegaard}.} \bibinfo{year}{2001}\natexlab{}.
\newblock \showarticletitle{Calculation of average PSNR differences between RD-curves}.
\newblock \bibinfo{journal}{\emph{ITU SG16 Doc. VCEG-M33}} (\bibinfo{year}{2001}).
\newblock


\bibitem[Bremner and Spence(2017)]%
        {bremner2017development}
\bibfield{author}{\bibinfo{person}{Andrew~J Bremner} {and} \bibinfo{person}{Charles Spence}.} \bibinfo{year}{2017}\natexlab{}.
\newblock \showarticletitle{The development of tactile perception}.
\newblock \bibinfo{journal}{\emph{Advances in child development and behavior}}  \bibinfo{volume}{52} (\bibinfo{year}{2017}), \bibinfo{pages}{227--268}.
\newblock


\bibitem[Cepri{\'a}-Bernal and P{\'e}rez-Gonz{\'a}lez(2021)]%
        {cepria2021dataset}
\bibfield{author}{\bibinfo{person}{Javier Cepri{\'a}-Bernal} {and} \bibinfo{person}{Antonio P{\'e}rez-Gonz{\'a}lez}.} \bibinfo{year}{2021}\natexlab{}.
\newblock \showarticletitle{Dataset of tactile signatures of the human right hand in twenty-one activities of daily living using a high spatial resolution pressure sensor}.
\newblock \bibinfo{journal}{\emph{Sensors}} \bibinfo{volume}{21}, \bibinfo{number}{8} (\bibinfo{year}{2021}), \bibinfo{pages}{2594}.
\newblock


\bibitem[Fung(2001)]%
        {clustersurvey}
\bibfield{author}{\bibinfo{person}{Glenn Fung}.} \bibinfo{year}{2001}\natexlab{}.
\newblock \showarticletitle{A comprehensive overview of basic clustering algorithms}.
\newblock  (\bibinfo{year}{2001}).
\newblock


\bibitem[Garcia-Garcia et~al\mbox{.}(2017)]%
        {garcia2017multi}
\bibfield{author}{\bibinfo{person}{Alberto Garcia-Garcia}, \bibinfo{person}{Sergio Orts-Escolano}, \bibinfo{person}{Sergiu Oprea}, \bibinfo{person}{Jose Garcia-Rodriguez}, \bibinfo{person}{Jorge Azorin-Lopez}, \bibinfo{person}{Marcelo Saval-Calvo}, {and} \bibinfo{person}{Miguel Cazorla}.} \bibinfo{year}{2017}\natexlab{}.
\newblock \showarticletitle{Multi-sensor 3D object dataset for object recognition with full pose estimation}.
\newblock \bibinfo{journal}{\emph{Neural Computing and Applications}}  \bibinfo{volume}{28} (\bibinfo{year}{2017}), \bibinfo{pages}{941--952}.
\newblock


\bibitem[Google(2010)]%
        {webp}
\bibfield{author}{\bibinfo{person}{Google}.} \bibinfo{year}{2010}\natexlab{}.
\newblock \bibinfo{title}{WebP Image Format}.
\newblock \bibinfo{howpublished}{\url{https://developers.google.com/speed/webp}}.
\newblock
\newblock
\shownote{Accessed: 2025-02-28}.


\bibitem[Hartigan and Wong(1979)]%
        {kmeans}
\bibfield{author}{\bibinfo{person}{John~A Hartigan} {and} \bibinfo{person}{Manchek~A Wong}.} \bibinfo{year}{1979}\natexlab{}.
\newblock \showarticletitle{Algorithm AS 136: A k-means clustering algorithm}.
\newblock \bibinfo{journal}{\emph{Journal of the royal statistical society. series c (applied statistics)}} \bibinfo{volume}{28}, \bibinfo{number}{1} (\bibinfo{year}{1979}), \bibinfo{pages}{100--108}.
\newblock


\bibitem[Hearst et~al\mbox{.}(1998)]%
        {svm}
\bibfield{author}{\bibinfo{person}{Marti~A. Hearst}, \bibinfo{person}{Susan~T Dumais}, \bibinfo{person}{Edgar Osuna}, \bibinfo{person}{John Platt}, {and} \bibinfo{person}{Bernhard Scholkopf}.} \bibinfo{year}{1998}\natexlab{}.
\newblock \showarticletitle{Support vector machines}.
\newblock \bibinfo{journal}{\emph{IEEE Intelligent Systems and their applications}} \bibinfo{volume}{13}, \bibinfo{number}{4} (\bibinfo{year}{1998}), \bibinfo{pages}{18--28}.
\newblock


\bibitem[Hollis et~al\mbox{.}(2016)]%
        {hollis2016compressed}
\bibfield{author}{\bibinfo{person}{Brayden Hollis}, \bibinfo{person}{Stacy Patterson}, {and} \bibinfo{person}{Jeff Trinkle}.} \bibinfo{year}{2016}\natexlab{}.
\newblock \showarticletitle{Compressed sensing for tactile skins}. In \bibinfo{booktitle}{\emph{2016 IEEE International Conference on Robotics and Automation (ICRA)}}. IEEE, \bibinfo{pages}{150--157}.
\newblock


\bibitem[ISO/IEC(2000)]%
        {jpeg2000}
\bibfield{author}{\bibinfo{person}{ISO/IEC}.} \bibinfo{year}{2000}\natexlab{}.
\newblock \bibinfo{title}{JPEG 2000 Image Coding System}.
\newblock \bibinfo{howpublished}{\url{https://www.jpeg.org/jpeg2000/}}.
\newblock
\newblock
\shownote{Accessed: 2025-02-28}.


\bibitem[(JVET)(2020)]%
        {vtm}
\bibfield{author}{\bibinfo{person}{Joint Video Experts~Team (JVET)}.} \bibinfo{year}{2020}\natexlab{}.
\newblock \bibinfo{title}{VVC Test Model (VTM)}.
\newblock \bibinfo{howpublished}{\url{https://jvet.hhi.fraunhofer.de/}}.
\newblock
\newblock
\shownote{Accessed: 2025-02-28}.


\bibitem[Kanitkar et~al\mbox{.}(2022)]%
        {9981562}
\bibfield{author}{\bibinfo{person}{Shubham Kanitkar}, \bibinfo{person}{Helen Jiang}, {and} \bibinfo{person}{Wenzhen Yuan}.} \bibinfo{year}{2022}\natexlab{}.
\newblock \showarticletitle{PoseIt: A Visual-Tactile Dataset of Holding Poses for Grasp Stability Analysis}. In \bibinfo{booktitle}{\emph{2022 IEEE/RSJ International Conference on Intelligent Robots and Systems (IROS)}}. \bibinfo{pages}{71--78}.
\newblock
\href{https://doi.org/10.1109/IROS47612.2022.9981562}{doi:\nolinkurl{10.1109/IROS47612.2022.9981562}}


\bibitem[Korhonen and You(2012)]%
        {psnr}
\bibfield{author}{\bibinfo{person}{Jari Korhonen} {and} \bibinfo{person}{Junyong You}.} \bibinfo{year}{2012}\natexlab{}.
\newblock \showarticletitle{Peak signal-to-noise ratio revisited: Is simple beautiful?}. In \bibinfo{booktitle}{\emph{2012 Fourth international workshop on quality of multimedia experience}}. IEEE, \bibinfo{pages}{37--38}.
\newblock


\bibitem[Lepora(2024)]%
        {lepora2024future}
\bibfield{author}{\bibinfo{person}{Nathan~F Lepora}.} \bibinfo{year}{2024}\natexlab{}.
\newblock \showarticletitle{The future lies in a pair of tactile hands}.
\newblock \bibinfo{journal}{\emph{Science Robotics}} \bibinfo{volume}{9}, \bibinfo{number}{91} (\bibinfo{year}{2024}), \bibinfo{pages}{eadq1501}.
\newblock


\bibitem[Li et~al\mbox{.}(2024)]%
        {10710144}
\bibfield{author}{\bibinfo{person}{Tong Li}, \bibinfo{person}{Yuhang Yan}, \bibinfo{person}{Chengshun Yu}, \bibinfo{person}{Jing An}, \bibinfo{person}{Yifan Wang}, \bibinfo{person}{Xiaojun Zhu}, {and} \bibinfo{person}{Gang Chen}.} \bibinfo{year}{2024}\natexlab{}.
\newblock \showarticletitle{VTG: A Visual-Tactile Dataset for Three-Finger Grasp}.
\newblock \bibinfo{journal}{\emph{IEEE Robotics and Automation Letters}} \bibinfo{volume}{9}, \bibinfo{number}{11} (\bibinfo{year}{2024}), \bibinfo{pages}{10684--10691}.
\newblock
\href{https://doi.org/10.1109/LRA.2024.3477168}{doi:\nolinkurl{10.1109/LRA.2024.3477168}}


\bibitem[Lima et~al\mbox{.}(2023)]%
        {lima2023multimodal}
\bibfield{author}{\bibinfo{person}{Bruno Monteiro~Rocha Lima}, \bibinfo{person}{Venkata Naga Sai~Siddhartha Danyamraju}, \bibinfo{person}{Thiago Eustaquio~Alves de Oliveira}, {and} \bibinfo{person}{Vinicius~Prado da Fonseca}.} \bibinfo{year}{2023}\natexlab{}.
\newblock \showarticletitle{A multimodal tactile dataset for dynamic texture classification}.
\newblock \bibinfo{journal}{\emph{Data in Brief}}  \bibinfo{volume}{50} (\bibinfo{year}{2023}), \bibinfo{pages}{109590}.
\newblock


\bibitem[Liu et~al\mbox{.}(2023)]%
        {tcm}
\bibfield{author}{\bibinfo{person}{Jinming Liu}, \bibinfo{person}{Heming Sun}, {and} \bibinfo{person}{Jiro Katto}.} \bibinfo{year}{2023}\natexlab{}.
\newblock \showarticletitle{Learned image compression with mixed transformer-cnn architectures}. In \bibinfo{booktitle}{\emph{Proceedings of the IEEE/CVF conference on computer vision and pattern recognition}}. \bibinfo{pages}{14388--14397}.
\newblock


\bibitem[loup Gailly and Adler(1992)]%
        {gzip}
\bibfield{author}{\bibinfo{person}{Jean loup Gailly} {and} \bibinfo{person}{Mark Adler}.} \bibinfo{year}{1992}\natexlab{}.
\newblock \bibinfo{title}{GNU Gzip}.
\newblock \bibinfo{howpublished}{\url{https://www.gnu.org/software/gzip/}}.
\newblock
\newblock
\shownote{Accessed: 2025-02-28}.


\bibitem[Lu and Weng(2007)]%
        {clssurvey}
\bibfield{author}{\bibinfo{person}{Dengsheng Lu} {and} \bibinfo{person}{Qihao Weng}.} \bibinfo{year}{2007}\natexlab{}.
\newblock \showarticletitle{A survey of image classification methods and techniques for improving classification performance}.
\newblock \bibinfo{journal}{\emph{International journal of Remote sensing}} \bibinfo{volume}{28}, \bibinfo{number}{5} (\bibinfo{year}{2007}), \bibinfo{pages}{823--870}.
\newblock


\bibitem[Lu et~al\mbox{.}(2025)]%
        {lu2025cross}
\bibfield{author}{\bibinfo{person}{Hang Lu}, \bibinfo{person}{Xinmeng Tan}, \bibinfo{person}{Mingkai Chen}, \bibinfo{person}{Zhe Zhang}, \bibinfo{person}{Xuguang Zhang}, \bibinfo{person}{Jianxin Chen}, \bibinfo{person}{Xin Wei}, {and} \bibinfo{person}{Tiesong Zhao}.} \bibinfo{year}{2025}\natexlab{}.
\newblock \showarticletitle{Cross-Modal Haptic Compression Inspired by Embodied AI for Haptic Communications}.
\newblock \bibinfo{journal}{\emph{IEEE Transactions on Multimedia}} (\bibinfo{year}{2025}).
\newblock


\bibitem[Ma et~al\mbox{.}(2019)]%
        {ma2019image}
\bibfield{author}{\bibinfo{person}{Siwei Ma}, \bibinfo{person}{Xinfeng Zhang}, \bibinfo{person}{Chuanmin Jia}, \bibinfo{person}{Zhenghui Zhao}, \bibinfo{person}{Shiqi Wang}, {and} \bibinfo{person}{Shanshe Wang}.} \bibinfo{year}{2019}\natexlab{}.
\newblock \showarticletitle{Image and video compression with neural networks: A review}.
\newblock \bibinfo{journal}{\emph{IEEE Transactions on Circuits and Systems for Video Technology}} \bibinfo{volume}{30}, \bibinfo{number}{6} (\bibinfo{year}{2019}), \bibinfo{pages}{1683--1698}.
\newblock


\bibitem[Mentzer et~al\mbox{.}(2019)]%
        {l3c}
\bibfield{author}{\bibinfo{person}{Fabian Mentzer}, \bibinfo{person}{Eirikur Agustsson}, \bibinfo{person}{Michael Tschannen}, \bibinfo{person}{Radu Timofte}, {and} \bibinfo{person}{Luc~Van Gool}.} \bibinfo{year}{2019}\natexlab{}.
\newblock \showarticletitle{Practical Full Resolution Learned Lossless Image Compression}. In \bibinfo{booktitle}{\emph{Proceedings of the IEEE/CVF Conference on Computer Vision and Pattern Recognition (CVPR)}}. \bibinfo{pages}{10629--10638}.
\newblock
\href{https://doi.org/10.1109/CVPR.2019.01089}{doi:\nolinkurl{10.1109/CVPR.2019.01089}}


\bibitem[Nguyen et~al\mbox{.}(2021)]%
        {scc-vtm}
\bibfield{author}{\bibinfo{person}{Tung Nguyen}, \bibinfo{person}{Xiaozhong Xu}, \bibinfo{person}{Felix Henry}, \bibinfo{person}{Ru-Ling Liao}, \bibinfo{person}{Mohammed~Golam Sarwer}, \bibinfo{person}{Marta Karczewicz}, \bibinfo{person}{Yung-Hsuan Chao}, \bibinfo{person}{Jizheng Xu}, \bibinfo{person}{Shan Liu}, \bibinfo{person}{Detlev Marpe}, {et~al\mbox{.}}} \bibinfo{year}{2021}\natexlab{}.
\newblock \showarticletitle{Overview of the screen content support in VVC: Applications, coding tools, and performance}.
\newblock \bibinfo{journal}{\emph{IEEE Transactions on Circuits and Systems for Video Technology}} \bibinfo{volume}{31}, \bibinfo{number}{10} (\bibinfo{year}{2021}), \bibinfo{pages}{3801--3817}.
\newblock


\bibitem[on~Video Coding (JCT-VC)(2013)]%
        {hm}
\bibfield{author}{\bibinfo{person}{Joint Collaborative~Team on Video Coding (JCT-VC)}.} \bibinfo{year}{2013}\natexlab{}.
\newblock \bibinfo{title}{HEVC Test Model (HM)}.
\newblock \bibinfo{howpublished}{\url{https://hevc.hhi.fraunhofer.de/}}.
\newblock
\newblock
\shownote{Accessed: 2025-02-28}.


\bibitem[PaXiniTech(2024)]%
        {paxiniDexH13}
\bibfield{author}{\bibinfo{person}{PaXiniTech}.} \bibinfo{year}{2024}\natexlab{}.
\newblock \bibinfo{title}{DexH13 Dexterous Hand}.
\newblock \bibinfo{howpublished}{Online}.
\newblock
\urldef\tempurl%
\url{https://mall.paxini.com/product/dex/66f27ea530cd11a8e9d1d5ff?skuId=66f27ec230cd11a8e9d1d610}
\showURL{%
\tempurl}
\newblock
\shownote{Accessed: 2025-02-28}.


\bibitem[Peterson(2009)]%
        {knn}
\bibfield{author}{\bibinfo{person}{Leif~E Peterson}.} \bibinfo{year}{2009}\natexlab{}.
\newblock \showarticletitle{K-nearest neighbor}.
\newblock \bibinfo{journal}{\emph{Scholarpedia}} \bibinfo{volume}{4}, \bibinfo{number}{2} (\bibinfo{year}{2009}), \bibinfo{pages}{1883}.
\newblock


\bibitem[Rigatti(2017)]%
        {randomforest}
\bibfield{author}{\bibinfo{person}{Steven~J Rigatti}.} \bibinfo{year}{2017}\natexlab{}.
\newblock \showarticletitle{Random forest}.
\newblock \bibinfo{journal}{\emph{Journal of Insurance Medicine}} \bibinfo{volume}{47}, \bibinfo{number}{1} (\bibinfo{year}{2017}), \bibinfo{pages}{31--39}.
\newblock


\bibitem[Seber and Lee(2012)]%
        {linregression}
\bibfield{author}{\bibinfo{person}{George~AF Seber} {and} \bibinfo{person}{Alan~J Lee}.} \bibinfo{year}{2012}\natexlab{}.
\newblock \bibinfo{booktitle}{\emph{Linear regression analysis}}.
\newblock \bibinfo{publisher}{John Wiley \& Sons}.
\newblock


\bibitem[Shao et~al\mbox{.}(2020)]%
        {shao2020compression}
\bibfield{author}{\bibinfo{person}{Yitian Shao}, \bibinfo{person}{Vincent Hayward}, {and} \bibinfo{person}{Yon Visell}.} \bibinfo{year}{2020}\natexlab{}.
\newblock \showarticletitle{Compression of dynamic tactile information in the human hand}.
\newblock \bibinfo{journal}{\emph{Science advances}} \bibinfo{volume}{6}, \bibinfo{number}{16} (\bibinfo{year}{2020}), \bibinfo{pages}{eaaz1158}.
\newblock


\bibitem[Slepyan et~al\mbox{.}(2024)]%
        {slepyan2024wavelet}
\bibfield{author}{\bibinfo{person}{Ariel Slepyan}, \bibinfo{person}{Michael Zakariaie}, \bibinfo{person}{Trac Tran}, {and} \bibinfo{person}{Nitish Thakor}.} \bibinfo{year}{2024}\natexlab{}.
\newblock \showarticletitle{Wavelet Transforms Significantly Sparsify and Compress Tactile Interactions}.
\newblock \bibinfo{journal}{\emph{Sensors}} \bibinfo{volume}{24}, \bibinfo{number}{13} (\bibinfo{year}{2024}), \bibinfo{pages}{4243}.
\newblock


\bibitem[Sneyers(2015)]%
        {flif}
\bibfield{author}{\bibinfo{person}{Jon Sneyers}.} \bibinfo{year}{2015}\natexlab{}.
\newblock \bibinfo{title}{FLIF - Free Lossless Image Format}.
\newblock \bibinfo{howpublished}{\url{https://flif.info/}}.
\newblock
\newblock
\shownote{Accessed: 2023-10-01}.


\bibitem[Steinbach et~al\mbox{.}(2018)]%
        {steinbach2018haptic}
\bibfield{author}{\bibinfo{person}{Eckehard Steinbach}, \bibinfo{person}{Matti Strese}, \bibinfo{person}{Mohamad Eid}, \bibinfo{person}{Xun Liu}, \bibinfo{person}{Amit Bhardwaj}, \bibinfo{person}{Qian Liu}, \bibinfo{person}{Mohammad Al-Ja’afreh}, \bibinfo{person}{Toktam Mahmoodi}, \bibinfo{person}{Rania Hassen}, \bibinfo{person}{Abdulmotaleb El~Saddik}, {et~al\mbox{.}}} \bibinfo{year}{2018}\natexlab{}.
\newblock \showarticletitle{Haptic codecs for the tactile internet}.
\newblock \bibinfo{journal}{\emph{Proc. IEEE}} \bibinfo{volume}{107}, \bibinfo{number}{2} (\bibinfo{year}{2018}), \bibinfo{pages}{447--470}.
\newblock


\bibitem[Sundaram et~al\mbox{.}(2019)]%
        {sundaram2019learning}
\bibfield{author}{\bibinfo{person}{Subramanian Sundaram}, \bibinfo{person}{Petr Kellnhofer}, \bibinfo{person}{Yunzhu Li}, \bibinfo{person}{Jun-Yan Zhu}, \bibinfo{person}{Antonio Torralba}, {and} \bibinfo{person}{Wojciech Matusik}.} \bibinfo{year}{2019}\natexlab{}.
\newblock \showarticletitle{Learning the signatures of the human grasp using a scalable tactile glove}.
\newblock \bibinfo{journal}{\emph{Nature}} \bibinfo{volume}{569}, \bibinfo{number}{7758} (\bibinfo{year}{2019}), \bibinfo{pages}{698--702}.
\newblock


\bibitem[Takahashi and Tan(2019)]%
        {takahashi2019deep}
\bibfield{author}{\bibinfo{person}{Kuniyuki Takahashi} {and} \bibinfo{person}{Jethro Tan}.} \bibinfo{year}{2019}\natexlab{}.
\newblock \showarticletitle{Deep visuo-tactile learning: Estimation of tactile properties from images}. In \bibinfo{booktitle}{\emph{2019 International Conference on Robotics and Automation (ICRA)}}. IEEE, \bibinfo{pages}{8951--8957}.
\newblock


\bibitem[Tatiya et~al\mbox{.}(2023)]%
        {tatiya2023transferring}
\bibfield{author}{\bibinfo{person}{Gyan Tatiya}, \bibinfo{person}{Jonathan Francis}, {and} \bibinfo{person}{Jivko Sinapov}.} \bibinfo{year}{2023}\natexlab{}.
\newblock \showarticletitle{Transferring implicit knowledge of non-visual object properties across heterogeneous robot morphologies}. In \bibinfo{booktitle}{\emph{2023 IEEE International Conference on Robotics and Automation (ICRA)}}. IEEE, \bibinfo{pages}{11315--11321}.
\newblock


\bibitem[Team(2021)]%
        {jpegxl}
\bibfield{author}{\bibinfo{person}{JPEG~XL Team}.} \bibinfo{year}{2021}\natexlab{}.
\newblock \bibinfo{title}{JPEG XL Image Coding System}.
\newblock \bibinfo{howpublished}{\url{https://jpeg.org/jpegxl/}}.
\newblock
\newblock
\shownote{Accessed: 2025-02-28}.


\bibitem[Van Den~Berg et~al\mbox{.}(2017)]%
        {8070953}
\bibfield{author}{\bibinfo{person}{Daniël Van Den~Berg}, \bibinfo{person}{Rebecca Glans}, \bibinfo{person}{Dorian De~Koning}, \bibinfo{person}{Fernando~A. Kuipers}, \bibinfo{person}{Jochem Lugtenburg}, \bibinfo{person}{Kurian Polachan}, \bibinfo{person}{Prabhakar~T. Venkata}, \bibinfo{person}{Chandramani Singh}, \bibinfo{person}{Belma Turkovic}, {and} \bibinfo{person}{Bryan Van~Wijk}.} \bibinfo{year}{2017}\natexlab{}.
\newblock \showarticletitle{Challenges in Haptic Communications Over the Tactile Internet}.
\newblock \bibinfo{journal}{\emph{IEEE Access}}  \bibinfo{volume}{5} (\bibinfo{year}{2017}), \bibinfo{pages}{23502--23518}.
\newblock
\href{https://doi.org/10.1109/ACCESS.2017.2764181}{doi:\nolinkurl{10.1109/ACCESS.2017.2764181}}


\bibitem[Van~der Maaten and Hinton(2008)]%
        {tsne}
\bibfield{author}{\bibinfo{person}{Laurens Van~der Maaten} {and} \bibinfo{person}{Geoffrey Hinton}.} \bibinfo{year}{2008}\natexlab{}.
\newblock \showarticletitle{Visualizing data using t-SNE.}
\newblock \bibinfo{journal}{\emph{Journal of machine learning research}} \bibinfo{volume}{9}, \bibinfo{number}{11} (\bibinfo{year}{2008}).
\newblock


\bibitem[Watkins et~al\mbox{.}(2018)]%
        {watkins2018image}
\bibfield{author}{\bibinfo{person}{Yijing Watkins}, \bibinfo{person}{Oleksandr Iaroshenko}, \bibinfo{person}{Mohammad Sayeh}, {and} \bibinfo{person}{Garrett Kenyon}.} \bibinfo{year}{2018}\natexlab{}.
\newblock \showarticletitle{Image compression: Sparse coding vs. bottleneck autoencoders}. In \bibinfo{booktitle}{\emph{2018 IEEE Southwest Symposium on Image Analysis and Interpretation (SSIAI)}}. IEEE, \bibinfo{pages}{17--20}.
\newblock


\bibitem[Xu et~al\mbox{.}(2015)]%
        {scc-hm}
\bibfield{author}{\bibinfo{person}{Jizheng Xu}, \bibinfo{person}{Rajan Joshi}, {and} \bibinfo{person}{Robert~A Cohen}.} \bibinfo{year}{2015}\natexlab{}.
\newblock \showarticletitle{Overview of the emerging HEVC screen content coding extension}.
\newblock \bibinfo{journal}{\emph{IEEE Transactions on Circuits and Systems for Video Technology}} \bibinfo{volume}{26}, \bibinfo{number}{1} (\bibinfo{year}{2015}), \bibinfo{pages}{50--62}.
\newblock


\bibitem[Xu and Liu(2021)]%
        {sccoverview}
\bibfield{author}{\bibinfo{person}{Xiaozhong Xu} {and} \bibinfo{person}{Shan Liu}.} \bibinfo{year}{2021}\natexlab{}.
\newblock \showarticletitle{Overview of screen content coding in recently developed video coding standards}.
\newblock \bibinfo{journal}{\emph{IEEE Transactions on Circuits and Systems for Video Technology}} \bibinfo{volume}{32}, \bibinfo{number}{2} (\bibinfo{year}{2021}), \bibinfo{pages}{839--852}.
\newblock


\bibitem[Xu et~al\mbox{.}(2024)]%
        {xu2024perception}
\bibfield{author}{\bibinfo{person}{Yiwen Xu}, \bibinfo{person}{Qingfeng Huang}, \bibinfo{person}{Quanfei Zheng}, \bibinfo{person}{Ying Fang}, {and} \bibinfo{person}{Tiesong Zhao}.} \bibinfo{year}{2024}\natexlab{}.
\newblock \showarticletitle{Perception-Based Prediction for Efficient Kinesthetic Coding}.
\newblock \bibinfo{journal}{\emph{IEEE Signal Processing Letters}} (\bibinfo{year}{2024}).
\newblock


\bibitem[Yang et~al\mbox{.}(2021)]%
        {yang2021probabilistic}
\bibfield{author}{\bibinfo{person}{Jun Yang}, \bibinfo{person}{Dong Li}, {and} \bibinfo{person}{Steven~L Waslander}.} \bibinfo{year}{2021}\natexlab{}.
\newblock \showarticletitle{Probabilistic multi-view fusion of active stereo depth maps for robotic bin-picking}.
\newblock \bibinfo{journal}{\emph{IEEE Robotics and Automation Letters}} \bibinfo{volume}{6}, \bibinfo{number}{3} (\bibinfo{year}{2021}), \bibinfo{pages}{4472--4479}.
\newblock


\bibitem[Yuan et~al\mbox{.}(2017)]%
        {yuan2017gelsight}
\bibfield{author}{\bibinfo{person}{Wenzhen Yuan}, \bibinfo{person}{Siyuan Dong}, {and} \bibinfo{person}{Edward~H Adelson}.} \bibinfo{year}{2017}\natexlab{}.
\newblock \showarticletitle{Gelsight: High-resolution robot tactile sensors for estimating geometry and force}.
\newblock \bibinfo{journal}{\emph{Sensors}} \bibinfo{volume}{17}, \bibinfo{number}{12} (\bibinfo{year}{2017}), \bibinfo{pages}{2762}.
\newblock


\bibitem[Zhao et~al\mbox{.}(2024)]%
        {zhao2024egopressure}
\bibfield{author}{\bibinfo{person}{Yiming Zhao}, \bibinfo{person}{Taein Kwon}, \bibinfo{person}{Paul Streli}, \bibinfo{person}{Marc Pollefeys}, {and} \bibinfo{person}{Christian Holz}.} \bibinfo{year}{2024}\natexlab{}.
\newblock \showarticletitle{EgoPressure: A Dataset for Hand Pressure and Pose Estimation in Egocentric Vision}.
\newblock \bibinfo{journal}{\emph{arXiv preprint arXiv:2409.02224}} (\bibinfo{year}{2024}).
\newblock


\bibitem[Ziv and Lempel(1977)]%
        {lz77}
\bibfield{author}{\bibinfo{person}{Jacob Ziv} {and} \bibinfo{person}{Abraham Lempel}.} \bibinfo{year}{1977}\natexlab{}.
\newblock \showarticletitle{A universal algorithm for sequential data compression}.
\newblock \bibinfo{journal}{\emph{IEEE Transactions on information theory}} \bibinfo{volume}{23}, \bibinfo{number}{3} (\bibinfo{year}{1977}), \bibinfo{pages}{337--343}.
\newblock


\end{thebibliography}

\end{document}